\documentclass{article}
\usepackage{spconf,amsmath,epsfig}

\usepackage{subfigure}
\usepackage{booktabs}
\usepackage{multirow}
\usepackage{tabularx}
\usepackage{hyperref}

\def\onedot{.}
\def\eg{\emph{e.g}\onedot} 
\def\ie{\emph{i.e}\onedot}

\def\etal{\emph{et al}\onedot}

\title{FlipReID: Closing the Gap between Training and Inference in \\Person Re-Identification}

\name{Xingyang Ni \quad Esa Rahtu}
\address{Tampere University, Finland}
\begin{document}

\maketitle

\begin{abstract}
Since neural networks are data-hungry, incorporating data augmentation in training is a widely adopted technique that enlarges datasets and improves generalization.
On the other hand, aggregating predictions of multiple augmented samples (\ie, test-time augmentation) could boost performance even further.
In the context of person re-identification models, it is common practice to extract embeddings for both the original images and their horizontally flipped variants.
The final representation is the mean of the aforementioned feature vectors.
However, such scheme results in a gap between training and inference, \ie, the mean feature vectors calculated in inference are not part of the training pipeline.
In this study, we devise the FlipReID structure with the flipping loss to address this issue.
More specifically, models using the FlipReID structure are trained on the original images and the flipped images simultaneously, and incorporating the flipping loss minimizes the mean squared error between feature vectors of corresponding image pairs.
Extensive experiments show that our method brings consistent improvements.
In particular, we set a new record for MSMT17 which is the largest person re-identification dataset.
The source code is available at \url{https://github.com/nixingyang/FlipReID}.
\end{abstract}


\section{Introduction}

The purpose of person re-identification is retrieving the person of interest across multiple cameras, based on either images or videos~\cite{ye2020deep}.
While large-scale datasets~\cite{zheng2015scalable,ristani2016performance,wei2018person} have been collected, academic research is progressing in three major directions: feature representation learning, deep metric learning, and ranking optimization.

Feature representation learning refers to developing strategies for feature construction~\cite{ye2020deep}.
Early works only extract a global representation for each person image~\cite{zheng2017person,luo2019bag}.
Later on, incorporating local features from body parts/regions has been proven beneficial~\cite{varior2016siamese,sun2018beyond}.
In addition, Lin \etal~\cite{lin2019improving} propose leveraging the auxiliary attributes, while Wei \etal~\cite{wei2018person} generate synthetic images to reduce the performance drop when training and testing on different datasets.

Deep metric learning studies objective functions which are utilized to optimize neural networks~\cite{ye2020deep}.
With the identity loss, a model classifies the identities, and each identity is a distinct class~\cite{zheng2017person,luo2019bag}.
By comparison, the verification loss exploits the pairwise relationship between samples.
Varior \etal~\cite{varior2016siamese} adopt the contrastive loss function for learning the embeddings, and Zheng \etal~\cite{zheng2017discriminatively} use the binary verification loss by feeding positive and negative pairs.
Lastly, the triplet loss is based on the assumption that the distance between the positive pair should be smaller than the negative pair~\cite{hermans2017defense}.

Ranking optimization is a post-processing step that refines the retrieved ranking list~\cite{ye2020deep}.
Liu \etal~\cite{liu2013pop} devise a method that requires only one negative feedback.
Wang \etal~\cite{wang2016human} propose a hybrid model that learns cumulatively from users' feedback, and it is scalable to large-scale gallery sets.
Since human interaction is time-consuming, Zhong \etal~\cite{zhong2017re} opt for a fully automated solution instead.
The pairwise distance between query and gallery samples is updated by comparing their k-reciprocal nearest neighbors.

\begin{figure}[t]
\begin{center}
\includegraphics[width=0.95\linewidth]{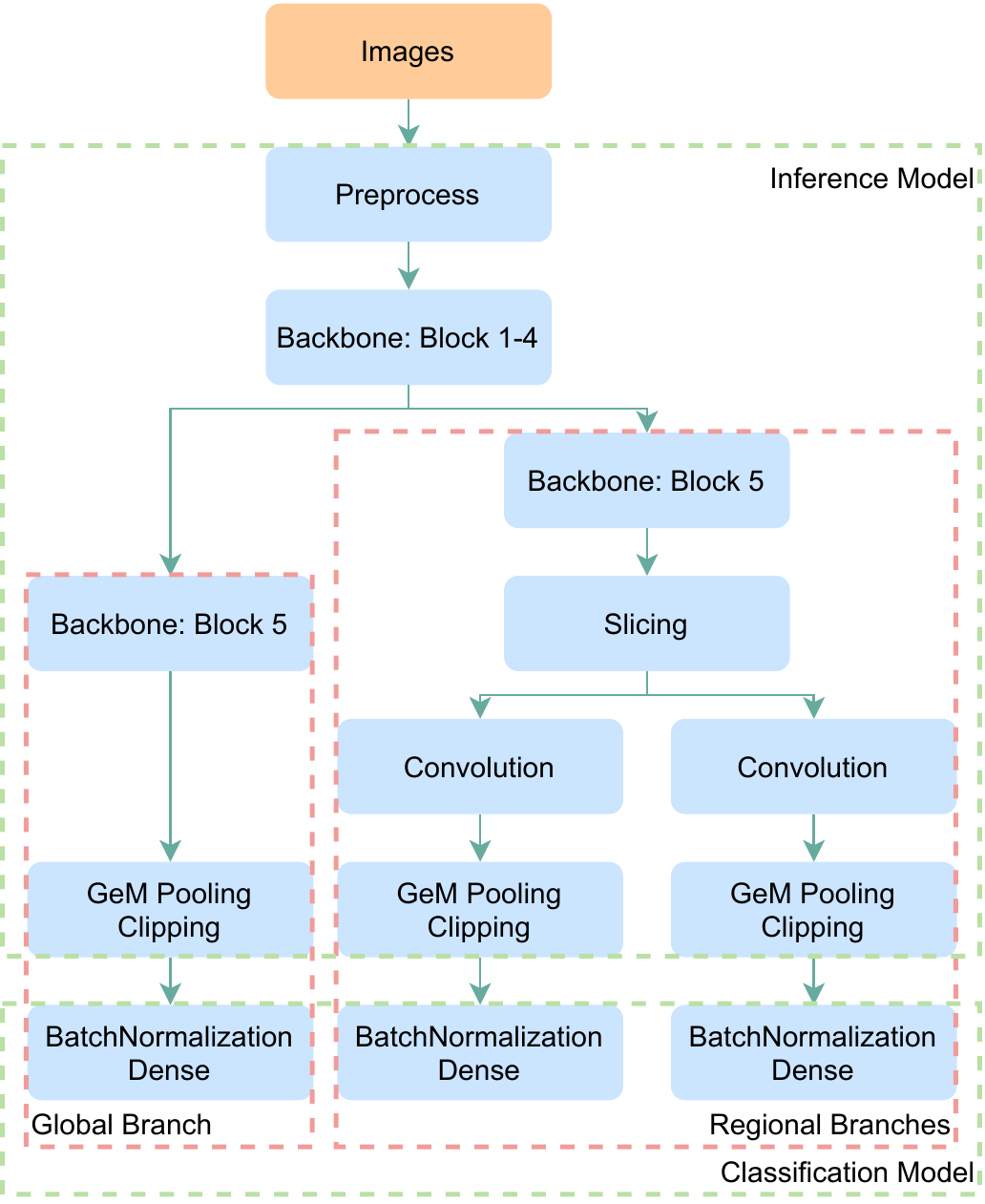}
\end{center}
\caption{
Overall structure of the baseline method.
}
\label{figure:baseline}
\end{figure}

Apart from person re-identification, data augmentation is the other subject of this study.
Transformations are applied on the samples while preserving the labels, thus avoiding the need for re-annotating~\cite{shorten2019survey}.
It is widely adopted in the training procedure to diversify samples, reduce overfitting and improve model robustness.
In the mixup~\cite{zhang2017mixup} work, models are trained on convex combinations of pairs of samples and their labels.
Cubuk \etal~\cite{cubuk2018autoaugment} propose a search algorithm that finds the best data augmentation policies for the task at hand.
The learned policies perform better than manually designed policies and are transferable between datasets.
Optionally, data augmentation can be utilized in the inference procedure as well, \ie, one could generate predictions of multiple augmented samples and aggregate them into a more accurate prediction.
Employing test-time augmentation improves performance at the cost of extra computations.

For image classification models such as AlexNet~\cite{krizhevsky2012imagenet} and ResNet~\cite{he2016deep}, a 10-crop testing method is applied.
Five patches are extracted from the original image, \ie, four corner patches and one center patch.
Another five patches are obtained from the horizontal reflection.
Since the output of the last dense layer with softmax activation represents the probabilities of classes, it is trivial to use the mean of these ten patches' predictions.

For person re-identification models such as MGN~\cite{wang2018MGN} and FastReID~\cite{he2020fastreid}, features are extracted from both the original images and their horizontally flipped variants.
Afterward, the mean of these feature vectors is used for evaluation.
Even though this practice brings performance improvements, it lacks a thorough explanation for aggregating feature vectors by calculating the mean.
More specifically, such mean feature vectors are not involved in optimizing the model, and calculating the mean in inference deviates from the training objectives.
As a consequence, there exists a gap between training and inference.

In this study, we examine test-time augmentation in person re-identification.
Our contribution is twofold:
\begin{itemize}
\itemsep0em
\item
We identify a largely neglected issue, \ie, utilizing the mean feature vectors of multiple augmented samples for evaluation results in suboptimal performance due to the gap between training and inference.
\item
We devise the FlipReID structure with the flipping loss.
For models using the FlipReID structure, the original images and the flipped images are both used for training.
Additionally, incorporating the flipping loss minimizes the mean squared error between feature vectors of corresponding image pairs.
The resulting method achieves state-of-the-art performance on popular person re-identification datasets.
\end{itemize}

\section{Proposed method}

\subsection{Baseline}

We leverage the method described in~\cite{ni2021adaptive} as the baseline since it achieves high performance among recent works in person re-identification.
Figure~\ref{figure:baseline} illustrates the overall structure of the baseline method, and its essential components are explained as follows.

\noindent\textbf{Backbone.}
Given images with pixel values in $\{0,\ldots,255\}$, the pre-processing layer scales pixels to $[0,1]$ and normalizes each channel.
The backbone model is an image classification model pre-trained on ImageNet~\cite{deng2009imagenet}, \eg, ResNet~\cite{he2016deep}, IBN-ResNet~\cite{pan2018two} and ResNeSt~\cite{zhang2020resnest}.
Due to its pyramidal architecture, the backbone model can be separated into consecutive blocks, and higher blocks refine feature maps extracted by lower blocks.

\noindent\textbf{Global Branch.}
Following the last block of backbone, the Generalized-Mean (GeM) pooling~\cite{radenovic2018fine} layer learns a trainable power parameter, and it generalizes the max and average operations.
The subsequent clipping layer performs element-wise value clipping, which introduces constraints on feature vectors.
Lastly, the batch normalization~\cite{ioffe2015batch} and dense layers produce the probabilities of classes.

\noindent\textbf{Regional Branches.}
The regional branches differ from the global branch in two aspects.
On the one hand, the slicing layer~\cite{sun2018beyond} divides the feature maps along the height axis.
For example, it outputs the upper and lower stripes in the case of using two partitions.
On the other hand, the following convolution layer reduces the number of channels so that the resulting feature vector is not excessively long.

\noindent\textbf{Objective Function.}
In the training procedure, the objective function is a weighted sum of batch hard triplet loss~\cite{hermans2017defense} and categorical cross-entropy loss.
The batch hard triplet loss utilizes the hardest positive and negative samples within the batch when forming the triplets, and it is applied to the clipping layer's output.
By contrast, the categorical cross-entropy loss measures the difference between the ground truth and predicted probability distributions, and it is applied to the dense layer's output.

\noindent\textbf{Sub-Models.}
In the inference procedure, the outputs of the clipping layers from both global branch and regional branches are concatenated to get the embedded feature vector.
From another perspective, the pipeline starting from the pre-processing layer to the clipping layers can be viewed as the inference model, while the remaining batch normalization and dense layers constitute the classification model.

\noindent\textbf{Mini-Batch.}
Due to the nature of the batch hard triplet loss~\cite{hermans2017defense}, each mini-batch comprises samples from both the same and different identities so that the positive and negative exemplars can be selected.
To alleviate the overfitting issue, random horizontal flipping, random grayscale patch replacement~\cite{gong2021effective} and random erasing~\cite{zhong2017random} are used as the data augmentation policies.

\begin{figure*}[t]
\begin{center}
\begin{tabular}{c|c|c|c}
\subfigure[][]{
\includegraphics[width=0.134\linewidth]{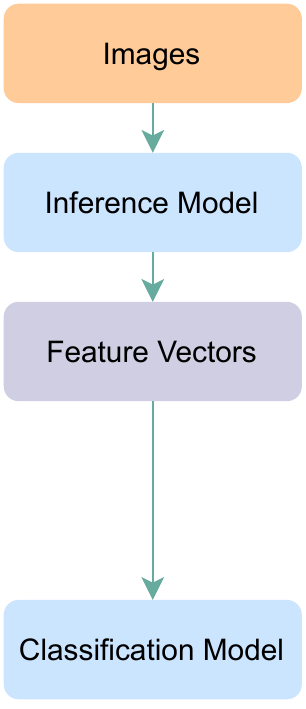}
\label{figure:training_baseline}
}&
\subfigure[][]{
\includegraphics[width=0.290\linewidth]{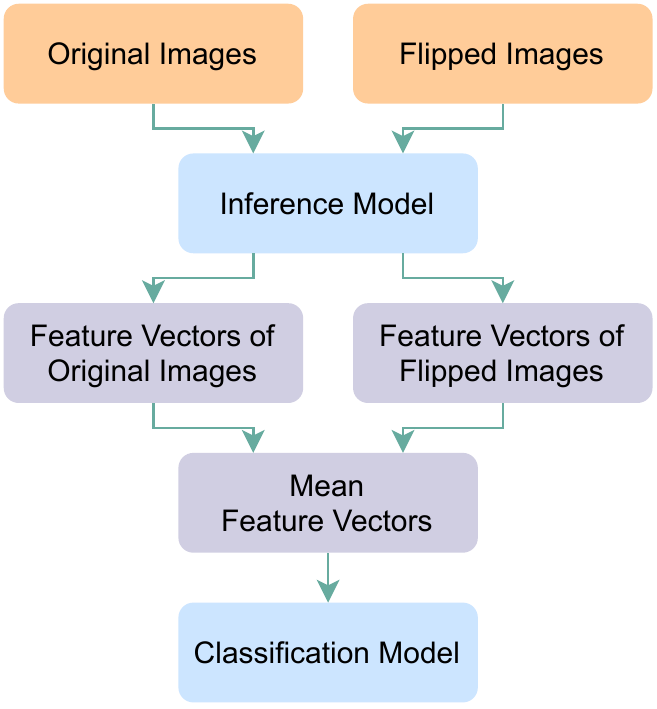}
\label{figure:training_FlipReID}
}&
\subfigure[][]{
\includegraphics[width=0.134\linewidth]{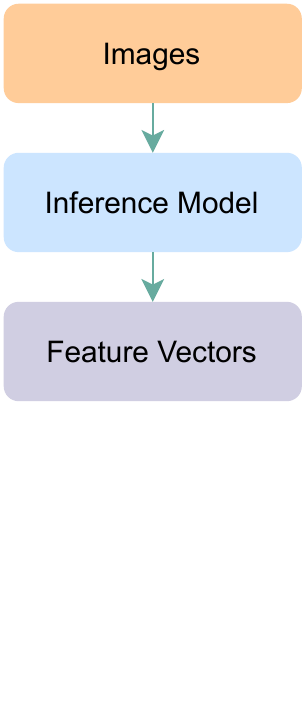}
\label{figure:inference_single}
}&
\subfigure[][]{
\includegraphics[width=0.290\linewidth]{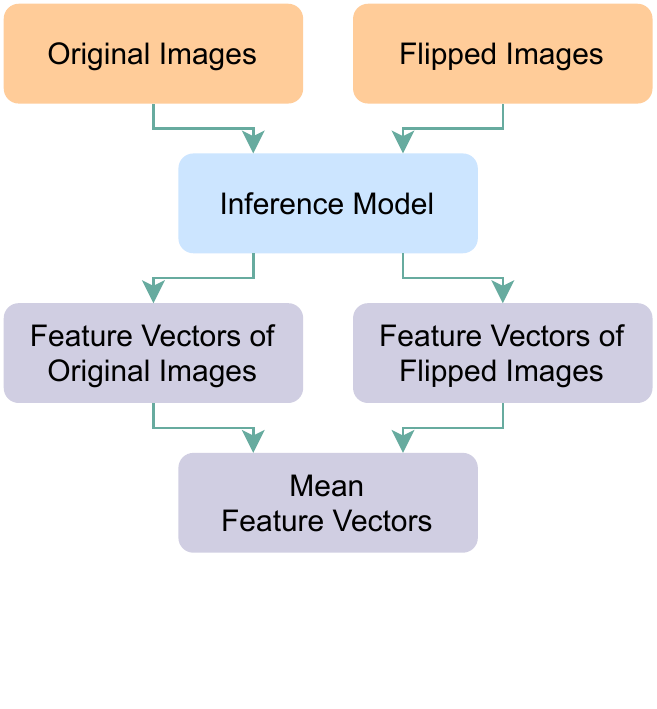}
\label{figure:inference_double}
}
\end{tabular}
\end{center}
\caption{
Overview of the training and inference procedures in different settings:
\subref{figure:training_baseline} training in the baseline method;
\subref{figure:training_FlipReID} training in the FlipReID method;
\subref{figure:inference_single} inference using single image;
\subref{figure:inference_double} inference using double images.
}
\label{figure:mix}
\end{figure*}

\subsection{FlipReID}

\noindent\textbf{Training.}
Figure~\ref{figure:training_baseline} visualizes a higher level of abstraction of Figure~\ref{figure:baseline}.
In the baseline method, random horizontal flipping is adopted for data augmentation.
However, only one image will be used for one sample at a time, \ie, either the original image or the flipped image.
By contrast, Figure~\ref{figure:training_FlipReID} shows the diagram of the FlipReID method.
Both the original images and the flipped images are feed into the inference model, and the classification model takes in the mean feature vectors.
As an optional loss term, we introduce the flipping loss, which calculates the mean squared error between the feature vectors of the original images and the flipped images.

\noindent\textbf{Inference.}
Figure~\ref{figure:inference_single} and~\ref{figure:inference_double} show the inference pipeline using single image and double images, respectively.
In Figure~\ref{figure:inference_single}, data augmentation is disabled, and feature vectors are extracted from the original images.
By comparison, horizontally flipped images are used in Figure~\ref{figure:inference_double}, and the final representation is the mean feature vectors of the original images and horizontal reflections.

\noindent\textbf{Options for Training and Inference.}
Figure~\ref{figure:training_baseline} and~\ref{figure:training_FlipReID} provide two choices for training, while Figure~\ref{figure:inference_single} and~\ref{figure:inference_double} offer two inference strategies.
This gives the following four options for training and inference:
\begin{itemize}
\itemsep0em
\item
Figure~\ref{figure:training_baseline}, \ref{figure:inference_single}:
Inference is consistent with training, and the classification model is discarded in inference.
\item
Figure~\ref{figure:training_baseline}, \ref{figure:inference_double}:
There exists a gap between training and inference, \ie,
the classification model is only trained on feature vectors of single image.
\item
Figure~\ref{figure:training_FlipReID}, \ref{figure:inference_single}:
There exists a gap between training and inference, \ie,
the classification model is only trained on feature vectors of double images.
\item
Figure~\ref{figure:training_FlipReID}, \ref{figure:inference_double}:
Inference is consistent with training, and the classification model is discarded in inference.
\end{itemize}

\section{Experiments}

\subsection{Datasets}

Experiments are carried out on three person re-identification datasets, \ie, Market-1501~\cite{zheng2015scalable}, DukeMTMC-reID~\cite{ristani2016performance} and MSMT17~\cite{wei2018person}.

\noindent\textbf{Market-1501.}
It consists of 32,217 images taken from 1,501 pedestrians.
Each pedestrian is captured by at least two cameras, and there are six cameras in total.

\noindent\textbf{DukeMTMC-reID.}
Eight cameras were deployed to collect the dataset.
The training set contains 702 pedestrians with 16,522 images, and the test set contains 1,110 pedestrians with 19,889 images.
Note that 408 pedestrians appear only in one camera, and those samples serve as distractors.

\noindent\textbf{MSMT17.}
Compared with the previous two datasets, the MSMT17 dataset is closer to real scenarios due to its diversity.
Collected by three indoor cameras and twelve outdoor cameras, it comprises 126,441 images from 4,101 pedestrians.
The ratio of training to test samples is set to 1:3 with the intention of limiting available training samples and therefore emphasizing effective training methods.

\subsection{Evaluation metrics}

Models are evaluated using mean Average Precision (mAP) and Cumulative Matching Characteristic (CMC) rank-k accuracy~\cite{zheng2015scalable}.
The mAP score is preferable to the CMC rank-k accuracy because the former metric considers both precision and recall while the latter metric does not report the performance on hard positive samples.
Furthermore, gallery samples taken from the same camera as the query sample are discarded during evaluation so that the metrics would emphasize the performance in the cross-camera setting.

\begin{table*}[t]
\renewcommand{\arraystretch}{0.7}
\caption{
Performance comparisons among existing studies, our baseline method, and our FlipReID method.
$\dag$: inference using single image (see Figure~\ref{figure:inference_single}).
$\ddag$: inference using double images (see Figure~\ref{figure:inference_double}).
$\S$: re-ranking~\cite{zhong2017re} is applied.
}
\label{table:performance_comparisons}
\centering
\begin{tabularx}{0.95\textwidth}{@{}lc*{6}{>{\centering\arraybackslash}X}@{}}
\toprule
\multirow{2}{*}{Method} & \multirow{2}{*}{Backbone} & \multicolumn{2}{c}{Market-1501} & \multicolumn{2}{c}{DukeMTMC-reID} & \multicolumn{2}{c}{MSMT17} \\
& & mAP & R1 & mAP & R1 & mAP & R1 \\
\midrule
PCB, ECCV 2018~\cite{sun2018beyond} & ResNet50 & 81.6 & 93.8 & 69.2 & 83.3 & - & - \\
BoT, CVPRW 2019~\cite{luo2019bag,he2020fastreid} & ResNet50 & 86.1 & 94.4 & 77.0 & 87.2 & 50.2 & 74.1 \\
SCSN, CVPR 2020~\cite{chen2020salience} & ResNet50 & 88.5 & 95.7 & 79.0 & 91.0 & 58.5 & 83.8 \\
GASM, ECCV 2020~\cite{he2020guided} & ResNet50 & 84.7 & 95.3 & 74.4 & 88.3 & 52.5 & 79.5 \\
AGW, TPAMI 2020~\cite{ye2020deep,he2020fastreid} & ResNet50 & 88.2 & 95.3 & 79.9 & 89.0 & 55.6 & 78.3 \\
FastReID, arXiv 2020~\cite{he2020fastreid} & ResNet50 & 88.2 & 95.4 & 79.8 & 89.6 & 59.9 & 83.3 \\
\midrule
1.1 Baseline\dag & \multirow{2}{*}{ResNet50} & 88.1\typeout{18210312} & 95.0 & 78.9\typeout{18210313} & 89.4 & 61.7\typeout{18210314} & 81.5 \\
1.2 Baseline\ddag & & 88.6\typeout{18219393} & 95.0 & 79.5\typeout{18219401} & 89.4 & 62.9\typeout{18219404} & 82.1 \\
\midrule
2.1 FlipReID (without flipping loss)\dag & \multirow{2}{*}{ResNet50} & 86.2\typeout{18211308} & 94.7 & 77.2\typeout{18211309} & 88.9 & 57.1\typeout{18211310} & 79.5 \\
2.2 FlipReID (without flipping loss)\ddag & & 88.5\typeout{18264277} & 95.5 & 79.8\typeout{18264322} & 90.2 & 64.3\typeout{18269144} & 83.6 \\
\midrule
3.1 FlipReID (with flipping loss)\dag & \multirow{3}{*}{ResNet50} & 87.6\typeout{18211304} & 95.2 & 78.9\typeout{18211305} & 89.1 & 61.4\typeout{18211306} & 81.9 \\
3.2 FlipReID (with flipping loss)\ddag & & 88.5\typeout{18264385} & 95.3 & 79.8\typeout{18264400} & 89.4 & 64.3\typeout{18264414} & 83.3 \\
3.3 FlipReID (with flipping loss)\ddag\S & & 94.6\typeout{18264427} & 96.0 & 90.9\typeout{18264433} & 92.5 & 79.5\typeout{18264438} & 86.3 \\
\midrule
1.1 Baseline\dag & \multirow{2}{*}{IBN-ResNet50} & 88.4\typeout{18210311} & 94.8 & 79.0\typeout{18210310} & 88.7 & 64.6\typeout{18210309} & 83.4 \\
1.2 Baseline\ddag & & 88.9\typeout{18219398} & 95.5 & 79.6\typeout{18219402} & 88.8 & 65.7\typeout{18219406} & 84.2 \\
\midrule
2.1 FlipReID (without flipping loss)\dag & \multirow{2}{*}{IBN-ResNet50} & 86.9\typeout{18210301} & 94.3 & 77.5\typeout{18210299} & 88.7 & 60.4\typeout{18210300} & 81.3 \\
2.2 FlipReID (without flipping loss)\ddag & & 88.7\typeout{18264302} & 94.8 & 79.7\typeout{18264335} & 89.4 & 66.2\typeout{18264353} & 84.4 \\
\midrule
3.1 FlipReID (with flipping loss)\dag & \multirow{3}{*}{IBN-ResNet50} & 87.9\typeout{18210215} & 94.7 & 78.8\typeout{18210200} & 88.9 & 63.2\typeout{18210199} & 83.1 \\
3.2 FlipReID (with flipping loss)\ddag & & 88.6\typeout{18264391} & 95.0 & 79.8\typeout{18264405} & 89.6 & 65.9\typeout{18264419} & 84.5 \\
3.3 FlipReID (with flipping loss)\ddag\S & & 94.1\typeout{18264429} & 95.4 & 89.8\typeout{18264436} & 91.8 & 80.2\typeout{18264439} & 87.3 \\
\midrule
1.1 Baseline\dag & \multirow{2}{*}{ResNeSt50} & 89.3\typeout{18210308} & 95.8 & 80.0\typeout{18210307} & 89.6 & 66.0\typeout{18210306} & 84.2 \\
1.2 Baseline\ddag & & 89.7\typeout{18219399} & 96.2 & 80.5\typeout{18219403} & 89.9 & 66.9\typeout{18219407} & 84.5 \\
\midrule
2.1 FlipReID (without flipping loss)\dag & \multirow{2}{*}{ResNeSt50} & 88.6\typeout{18210071} & 95.2 & 79.9\typeout{18210070} & 90.0 & 64.6\typeout{18210027} & 83.9 \\
2.2 FlipReID (without flipping loss)\ddag & & 89.6\typeout{18247746} & 95.7 & 81.2\typeout{18247741} & 90.7 & 67.6\typeout{18264365} & 85.3 \\
\midrule
3.1 FlipReID (with flipping loss)\dag & \multirow{3}{*}{ResNeSt50} & 88.9\typeout{18209984} & 95.2 & 80.7\typeout{18209982} & 90.5 & 66.0\typeout{18209952} & 84.6 \\
3.2 FlipReID (with flipping loss)\ddag & & 89.6\typeout{18247701} & 95.5 & 81.5\typeout{18247682} & 90.9 & 68.0\typeout{18234241} & 85.6 \\
3.3 FlipReID (with flipping loss)\ddag\S & & 94.7\typeout{18247728} & 95.8 & 90.7\typeout{18247730} & 93.0 & 81.3\typeout{18234242} & 87.5 \\
\bottomrule
\end{tabularx}
\end{table*}

\subsection{Analysis of results}

Table~\ref{table:performance_comparisons} compares the mAP scores and the rank-1 accuracies of existing studies and our methods.

\noindent\textbf{Datasets.}
Limited by the number and quality of samples, the Market-1501~\cite{zheng2015scalable} and DukeMTMC-reID~\cite{ristani2016performance} datasets are saturated, and scores reported on these two datasets may not be indicative~\cite{ni2021adaptive}.
For example, the FastReID~\cite{he2020fastreid} method surpasses the AGW~\cite{ye2020deep} method by a large margin on MSMT17~\cite{wei2018person}, while the scores on Market-1501 and DukeMTMC-reID are close.
In the remainder of this study, we mainly compare the mAP scores on MSMT17.

\noindent\textbf{Existing Studies versus Baseline.}
Among methods built on the ResNet50~\cite{he2016deep} backbone, it is evident that the baseline method performs better than existing studies on MSMT17.
This makes a good starting point since the FlipReID method is an extension of the baseline method.

\noindent\textbf{Single Image versus Double Images.}
Independent of how the training is performed, utilizing data augmentation in inference always boosts performance.
The notable downside of test-time augmentation is the extra computations which may pose a constraint for real-time applications, in which execution speed is crucial.

\noindent\textbf{Inconsistency between Figure~\ref{figure:training_baseline} and~\ref{figure:inference_double}, \ie, entries starting with 1.2.}
If data augmentation is enabled in inference, choosing the baseline scheme during training leads to suboptimal performance because the classification model is only trained on feature vectors of single image.
Such gap can be solved by switching to the FlipReID method, and noticeable improvement can be observed.

\noindent\textbf{Inconsistency between Figure~\ref{figure:training_FlipReID} and~\ref{figure:inference_single}, \ie, entries starting with 2.1 or 3.1.}
If data augmentation is disabled in inference, selecting the FlipReID mechanism during training results in inferior performance since the classification model is only trained on feature vectors of double images.
Adding the flipping loss is an effective approach to suppress this problem.
On the one hand, the resulting method is on par with the baseline method when using single image in inference.
On the other hand, the flipping loss does not degrade performance when using double images in inference.

\noindent\textbf{Backbone and Re-Ranking.}
In addition to ResNet~\cite{he2016deep}, experiments have been conducted with IBN-ResNet~\cite{pan2018two} and ResNeSt~\cite{zhang2020resnest}.
The latter two backbone models improve performance.
Moreover, adding re-ranking~\cite{zhong2017re} as a post-processing step introduces significant improvements.

\section{Conclusion}

In this study, we recognize and investigate the gap between training and inference in person re-identification.
Prior works typically use the mean of feature vectors extracted from the original images and their horizontally flipped variants in inference.
However, such mean feature vectors are not present when optimizing the model in training.
In order to close the gap, we propose to utilize the FlipReID structure with the flipping loss.
On the one hand, both the original images and the flipped images are feed into models with the FlipReID structure.
On the other hand, incorporating the flipping loss minimizes the mean squared error between feature vectors of corresponding image pairs.
Using the proposed method, models work as expected regardless of whether test-time augmentation is enabled or not, and the inconsistency issue is solved.
An extension of this study is to design a module that learns to aggregate feature vectors from multiple sources, rather than calculating the mean.

\bibliographystyle{IEEEbib}
\bibliography{nixingyang_references}

\end{document}